\documentclass[lettersize,journal]{IEEEtran}
\usepackage{amsmath,amsfonts}
\usepackage{algorithmic}
\usepackage{algorithm}
\usepackage{array}
\usepackage[caption=false,font=normalsize,labelfont=sf,textfont=sf]{subfig}
\usepackage{textcomp}
\usepackage{stfloats}
\usepackage{url}
\usepackage{verbatim}
\usepackage{graphicx}
\usepackage{cite}
\usepackage{multirow}
\usepackage{xcolor}
\usepackage{caption}
\usepackage{subcaption}
\usepackage[colorlinks,linkcolor=blue]{hyperref}

\begin{document}

\title{Weakly-Supervised 3D Visual Grounding based on Visual Language Alignment}

\author{Xiaoxu~Xu$^\dagger$,
			Yitian~Yuan$^\dagger$,
			Qiudan~Zhang,~\IEEEmembership{Member,~IEEE,}
                Wenhui~Wu,~\IEEEmembership{Member,~IEEE}
			Zequn~Jie,
			Lin~Ma,~\IEEEmembership{Member,~IEEE}
   			Xu~Wang,~\IEEEmembership{Member,~IEEE}

    \thanks{This work was supported in part by the National Natural Science Foundation of China (Grant 62371310, 62376162), in part by the Guangdong Basic and Applied Basic Research Foundation under Grant 2023A1515011236, and in part by the Stable Support Project of Shenzhen (Project No.20231122122722001). \emph{(Corresponding Author: Dr. Xu Wang)}
				
    Xiaoxu Xu, Qiudan Zhang and Xu Wang are with the College of Computer Science and Software Engineering, Shenzhen University, Shenzhen, 518060, China. Email: (xuxiaoxu68@163.com; qiudanzhang@szu.edu.cn; wangxu@szu.edu.cn).
				
    Yitian Yuan, Zequn Jie, and Lin Ma are with Meituan Inc., China. Email: (yuanyitian@foxmail.com; zequn.nus@gmail.com; forest.linma@gmail.com).

    Wenhui Wu is with College of Electronics and Information Engineering, Shenzhen University, China, and also with the Guangdong Key Laboratory of Intelligent Information Processing, Shenzhen, China, (email: wuwenhui@szu.edu.cn).

    {$^{\dagger}$These authors contributed equally to this work.}
				
}}

\markboth{submitted to IEEE Transactions on Multimedia, May~2024}%
{Shell \MakeLowercase{\textit{\textit{et al}.}}: A Sample Article Using IEEEtran.cls for IEEE Journals}


\maketitle

\begin{abstract}
Learning to ground natural language queries to target objects or regions in 3D point clouds is quite essential for 3D scene understanding. Nevertheless, existing 3D visual grounding approaches require a substantial number of bounding box annotations for text queries, which is time-consuming and labor-intensive to obtain. In this paper, we propose \textbf{3D-VLA}, a weakly supervised approach for \textbf{3D} visual grounding based on \textbf{V}isual \textbf{L}anguage \textbf{A}lignment. Our 3D-VLA exploits the superior ability of current large-scale vision-language models (VLMs) on aligning the semantics between texts and 2D images, as well as the naturally existing correspondences between 2D images and 3D point clouds, and thus implicitly constructs correspondences between texts and 3D point clouds with no need for fine-grained box annotations in the training procedure. During the inference stage, the learned text-3D correspondence will help us ground the text queries to the 3D target objects even without 2D images. To the best of our knowledge, this is the first work to investigate 3D visual grounding in a weakly supervised manner by involving large scale vision-language models, and extensive experiments on ReferIt3D and ScanRefer datasets demonstrate that our 3D-VLA achieves comparable and even superior results over the fully supervised methods.The code will be available at \href{https://github.com/xuxiaoxxxx/3D-VLA}{https://github.com/xuxiaoxxxx/3D-VLA}.
\end{abstract}


\section{Introduction}
\begin{figure}[t]
\centering
\includegraphics[width=1\columnwidth]{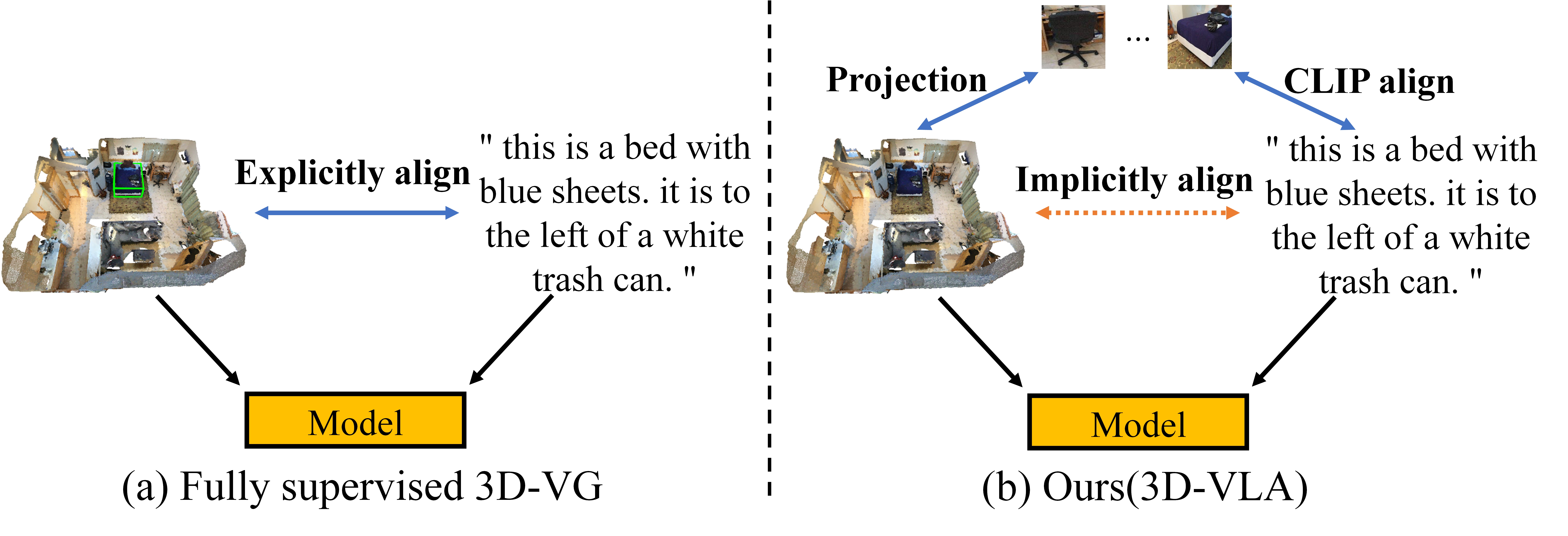} 
\caption{\small The comparison of fully supervised and our proposed weakly supervised 3D visual grounding. Our method leverages natural 3D-2D correspondence from geometric camera calibration and 2D-text correspondence from large-scale vision-language models to implicitly align texts and 3D point clouds.}
\label{fig:weaklyExample}
\end{figure}

3D visual grounding, which aims to precisely identify target objects in a 3D scene with the corresponding natural language queries, has gained considerable attention over the past few years~\cite{achlioptas2020referit3d, chen2020scanrefer, yuan2021instancerefer,chen2022d,yang2021sat,chen2022ham}. Previous works~\cite{he2021transrefer3d, zhao20213dvg,huang2021text,feng2021free,9830073} mainly explore fully supervised solutions for 3D visual grounding, as shown in Fig.\ref{fig:weaklyExample}~(a), the 3D bounding box for the text query is provided during the training procedure, which helps the model to establish the explicit alignment between the two modalities. However, annotating dense object-sentence in point clouds is labor-intensive and expensive, therefore it hinders large scale datasets collection, and further influences the model capability of 3D visual grounding.

To solve the above challenge, a natural way is to investigate 3D visual grounding in a weakly supervised manner that does not need dense object-sentence annotations. Such idea has been explored in 2D visual grounding, which mainly focus on establishing semantic correspondences between 2D image and text descriptions~\cite{datta2019align2ground, gupta2020contrastive, liu2021relation,10269126}. 
However, different from 2D images, 3D point clouds inherently provide essential geometric information and surface context with a higher level of complexity and a larger spatial scale, and bring new challenges to effectively learning the matching relationships between 3D point clouds and texts.

Wang \textit{et al}.~\cite{wang2023distilling} utilize a coarse-to-fine matching method with contrastive learning to identify top-$k$ candidate proposals, followed by text reconstruction loss for supervision. However, the low quality of candidate proposals selected in the first stage, coupled with reconstruction losses supervised solely by text embeddings, results in a alignment between the 3D and text relationships that is inadequate and not as expectation.
Therefore how to correlate texts and 3D point clouds is still a big challenge.

We can also notice that, the correspondences between 3D point clouds and 2D images can be easily obtained by geometric camera calibration with intrinsic and extrinsic parameters. At the same time, we can also note that the current pre-trained large-scale vision-language models (VLMs) such as CLIP~\cite{radford2021learning}, ViLT~\cite{kim2021vilt}, VLMO~\cite{bao2022vlmo} have been greatly developed. Using massive text-image pairs for model training, VLMs are able to establish precise semantic matching relationships between natural languages and 2D images, and have achieved good results in various downstream tasks such as image classification~\cite{chen2021crossvit,zhang2020deepemd}, visual question answering~\cite{jiang2020defense}, and image captioning~\cite{huang2019attention}. So, as shown in Fig.\ref{fig:weaklyExample}~(b), why don't we take 2D images as a bridge, leveraging the correspondences between point clouds and images, images and natural languages, to implicitly build matching relationships between point clouds and natural languages?

To this end, we present a novel weakly supervised method \textbf{3D-VLA}, which explores the \textbf{3D} visual grounding based on the \textbf{V}isual \textbf{L}anguage \textbf{A}lignment while without the need of 3D bounding box annotations. Specifically, as shown in Fig.~\ref{fig2}, in the training stage, our proposed 3D-VLA possesses a text module, a 2D module, and a 3D module. We first extract 3D proposal candidates from the point cloud scene and project these proposals to 2D image regions through geometric camera calibration, and then we utilize a frozen CLIP model to get the embeddings of the text query and 2D image regions with its text encoder and image encoder, respectively. The correspondences between the text query and 2D image regions can thus be measured through their CLIP embeddings. We leverage contrastive learning to optimize the 3D encoder in the 3D module by making the learned 3D embeddings comparable to the text and 2D CLIP embeddings. If a 2D image region and a 3D proposal are matched in pairs, their embeddings should be pulled closer, otherwise they should be pushed further apart. 

Ideally, if the 3D embedding of a proposal candidate is learned well enough, it can be directly compared with the text query embedding by the similarity measurement to judge whether it is the target proposal. 
However, we observe that relying solely on implicit contrastive learning is unreliable, as the pretrained data of VLMs is general and lacks specialized knowledge for indoor point cloud scenes. Indoor environments present greater complexity, characterized by higher object density and intricate spatial relationships, making accurate visual grounding when using only VLMs and contrastive learning methods. 
Therefore, we propose to alleviate this problem by introducing multi-modal adaption through task-aware classification. As shown in Fig.~\ref{fig2}, we first add three adapters to transfer the text, 2D and 3D embeddings to another embedding space, and then the 2D and 3D classification are realized by comparing the adapted region/proposal embeddings to the text embeddings of the category labels in the dataset. For the query, we directly apply a text classifier on its adapted query embedding, thus obtaining its distribution on the category labels. By introducing the task-aware classification signal of 3D visual grounding in the indoor point cloud scene, we can further align the semantic relationships among texts, 2D images and 3D point clouds specialized for 3D visual grounding.

In the inference stage, as shown in Fig.~\ref{fig3}, we can completely ignore the 2D image module and directly compare the learned 3D point cloud embeddings and text embeddings to determine the target proposal. At the same time, we can also use the  classification results of text and 3D objects to filter out some confusing and unreliable predictions. In summary, the main contributions of this paper are as follows:

\begin{itemize}
    \item We propose a weakly supervised method 3D-VLA for 3D visual grounding, which takes 2D images as a bridge, and leverages natural 3D-2D correspondence from geometric camera calibration and 2D-text correspondence from large-scale vision-language models to implicitly establish the semantic relationships between texts and 3D point clouds. 
    
    \item Our 3D-VLA utilizes contrastive learning to get 3D proposal embeddings that can basically align with the 2D and text embeddings from VLMs, and the introduced multi-modal adaption through task-aware classification also guides the learned embeddings to better support 3D visual grounding.

    \item Extensive experiments are conducted on two public datasets, and the experimental results demonstrate that our proposed 3D-VLA can achieve not only the state-of-the-art performances in the weakly supervised setting but also comparable and even superior results over the fully supervised methods. Our 3D-VLA and its results provide valuable insights to improve further research of weakly supervised 3D visual grounding.
    
    

    
\end{itemize}

\section{Related Work}

\subsection{Weakly Supervised Visual Grounding on Images} In contrast to the traditional supervised 2D visual grounding~\cite{deng2018visual,deng2021transvg,yang2022improving}, the weakly supervised setting focuses on learning the fine-grained correspondence between regions and phrases without relying on target bounding box annotations. Weakly supervised visual grounding on images is typically treated as a Multiple Instance Learning (MIL)~\cite{maron1997framework}  problem. In recent studies, a general approach for weakly supervised visual grounding~\cite{rohrbach2016grounding, chen2018knowledge, liu2019adaptive,wang-etal-2020-maf,dou2021improving,qu2023weakly} involves a hypothesis-and-matching strategy. Initially, a set of region proposals is generated from an image using an external object detector~\cite{salvador2016faster}. Then the model calculates the image-sentence matching scores and use the ground-truth image-sentence links to supervise these scores. For example, Chen \textit{et al}.~\cite{chen2018knowledge} leveraged pre-trained deep models and proposed to enforce visual and language consistency. InfoGround~\cite{gupta2020contrastive} improves the contrastive learning objective function to optimize image-sentence scores. Zhao \textit{et al}.~\cite{zhao2018weakly} jointly learns to propose object regions and matches the regions to phrases. Wang \textit{et al}.~\cite{wang2021improving} leverage the pre-trained image object detector to get the regions and their pseudo category labels, distilling knowledge from pseudo labels to align the region-phrase. 

However, there exit some problems that we cannot directly apply the method of 2D weakly-supervised visual grounding on the 3D weakly-supervised visual grounding task. Firstly, 3D point clouds inherently provide essential geometric information and surface context with a higher level of complexity and a larger spatial scale. For the 3D weakly-supervised visual grounding, there exit numerous different objects in a single 3D scene compared to the image visual grounding task, which makes the task more difficult. Secondly, while the objective of image grounding is to pinpoint objects corresponding to all phrases in the sentence, 3D visual grounding involves the identification of a solitary target object. This mandates a more profound and thorough comprehension of the semantic information conveyed by the sentence, extending beyond a mere focus on its individual phrases.

\subsection{3D Visual Grounding} The goal of 3D visual grounding is to find a matched 3D proposal described by the input text query and does not care which category it belongs to. The primary benchmark datasets for 3D visual grounding include ReferIt3D~\cite{achlioptas2020referit3d} and ScanRefer~\cite{chen2020scanrefer}, both of which are based on the ScanNet~\cite{dai2017scannet}. 
Previously, most approaches~\cite{huang2022multi, bakr2022look} adopt a two-stage pipeline. In the first stage, they employ a 3D object detector to generate object proposals. In the second stage, they search for the target proposal that best matches the given query. For instance, InstanceRefer~\cite{yuan2021instancerefer} predicts the target category from the language descriptions using a simple language classification model and jointly attributes, local relations and global localization aspects to select the most relevant instance.  Semantic-Assisted Training~\cite{yang2021sat} use the 2D semantic to help 3D visual grounding task during training but does not require 2D inputs during inference. Considering 3D scenes can freely rotate to different views and affect the position encoding, MVT~\cite{huang2022multi} proposes the Multi-View Transformer structure to fusion 3D scenes embeddings of different views.

However, owing to proposals generated in the first stage is of low quality, the performances of those models are limited. To address the issue of imprecise object proposals generated in the first stage, some one-stage pipeline~\cite{jain2022bottom, luo20223d, wu2023eda} are introduced. For example, 3D-SPS~\cite{luo20223d} directly performs 3D visual grounding at a single stage and treats 3D visual grounding task as a keypoint selection problem to find the most target-related keypoints. In order to well align visual-language feature, Wu \textit{et al}.~\cite{wu2023eda} propose a text decoupling module to parse language description into multiple semantic components.

Those methods mentioned above are all fully supervised, which need much expensive bounding box annotations. To overcome this shortcoming, Wang \textit{et al}.~\cite{wang2023distilling} adopt a two-stage coarse-to-fine semantic matching approach. In the first stage, they use contrastive learning to align 3D object and sentence query features, selecting top-$k$ object candidates based on whether the object-query pairs within the same scene are positive or negative. In the second stage, a semantic reconstruction module is introduced to compute the fine-grained semantic similarity between 3D objects and the sentence query, selecting the target object with the
lowest reconstruction loss from the candidates. However, in the one hand, in large 3D scenes with many objects, this two-stage matching process struggles to achieve precise alignment, especially in complex environments where object features overlap. In the other hand, The reconstruction loss is supervised solely by the text embeddings, which are too weak to reliably guide the model in selecting the best-matching object. These reasons cause that the relationship it build is not well-aligned and the performance of it is not as expectation. Therefore, how to build the well-aligned relationship between 3D point cloud and text is still a problem. 


\begin{figure*}[!h]
\centering
\includegraphics[width=1\textwidth]{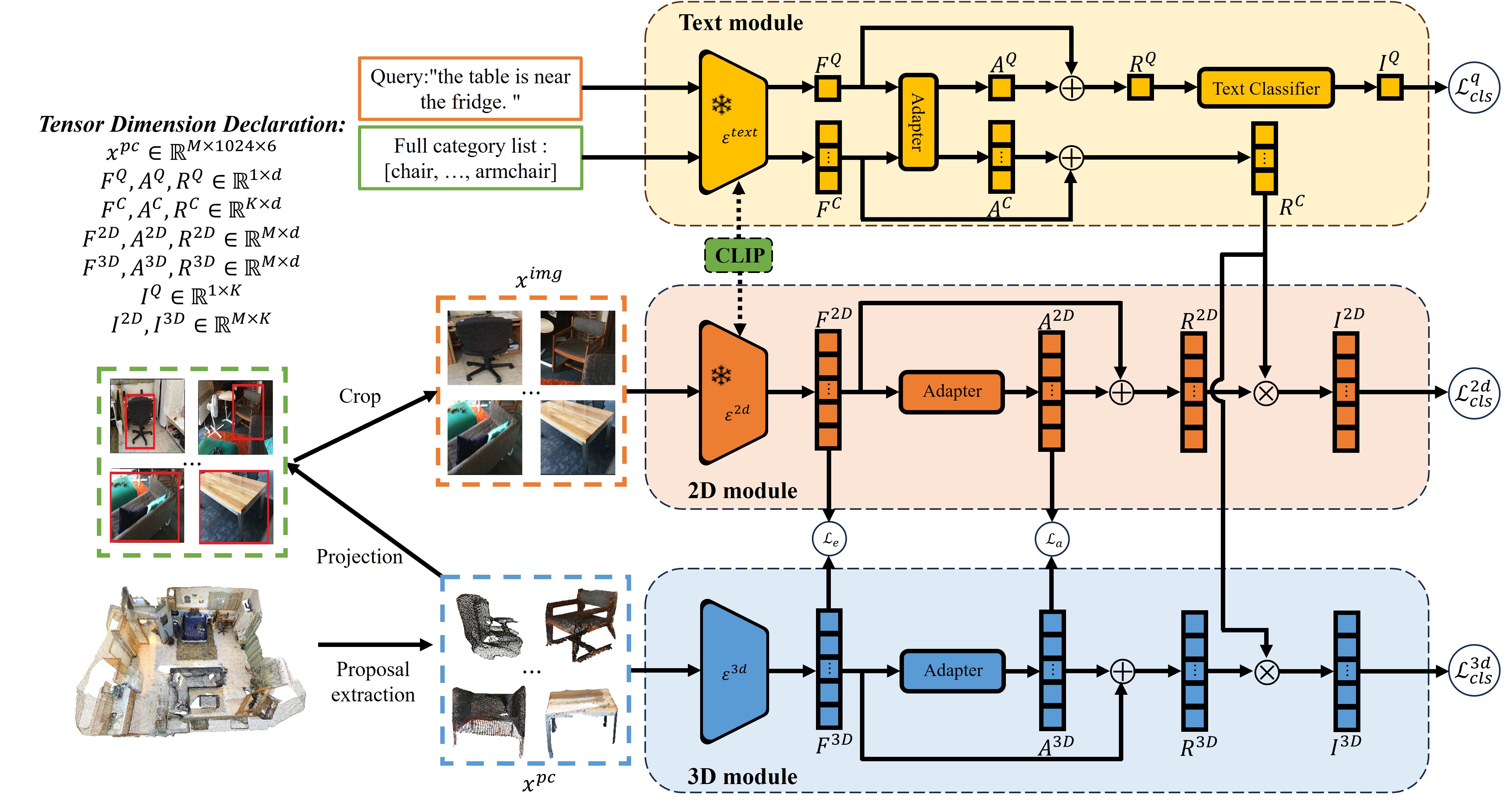} 
\caption{\small The training procedure of our proposed 3D-VLA. We first exact 3D proposal candidates $x^{pc}$ from the point cloud scene and use geometric camera calibration to project them to 2D image regions $x^{img}$. Then we leverage the text encoder $\varepsilon^{text}$ of CLIP to get embedding of the text query $F^Q$ and embedding of the category labels $F^C$, and leverage the 2D image encoder $\varepsilon^{2d}$ of CLIP to get embeddings of 2D image regions $F^{2D}$. It is important to note that we freeze the whole CLIP model during training. Meanwhile, we use 3D encoder $\varepsilon^{3d}$ to encode the 3D proposal candidates and get their 3D embeddings $F^{3D}$. Three adapters are further introduced to transfer the $F^C$, $F^{Q}$, $F^{2D}$, $F^{3D}$ to a new embedding space for coarse-grained classification in the indoor scene domain. We use contrastive learning to align the 2D CLIP embedding $F^{2D}$ and the encoded 3D embedding $F^{3D}$, and also align their corresponding adapted embeddings $A^{2D}$ and $A^{3D}$. The classification loss $\mathcal{L}_{cls}^{q}$, $\mathcal{L}_{cls}^{2d}$, $\mathcal{L}_{cls}^{3d}$ and the contrastive loss $\mathcal{L}_{e}$ and $\mathcal{L}_{a}$ will be integrated to train the overall model. } 
\label{fig2}
\end{figure*}

\subsection{2D Vision-Language Models} 
Exploring the interaction between vision and language is a core research topic in artificial intelligence. Vision-language models~\cite{kim2021vilt, bao2022vlmo, radford2021learning, lu2019vilbert, jia2021scaling,10017364} aim to leverage the text semantic to help some vision tasks. Among them, the Contrastive Language-Image Pretraining (CLIP)~\cite{radford2021learning} is most popular. It consists of an image encoder and a text encoder. Given a batch of image and text pairs, the CLIP model learns the embedding to measure the similarity between image and text. Owing to the well-aligned relationship between 2D image and text, CLIP shows great success and potential on many vision tasks in a zero-shot setting.

\subsection{3D Scene Understanding with 2D Semantics} Research on 3D tasks involves exploring how 2D image semantics can be integrated to provide assistance. These approaches typically utilise internal and external camera references to project 2D information into 3D space, thereby effectively aiding various tasks in the 3D domain.

However, in previous studies, the usage of 2D image semantics as additional inputs to 3D tasks necessitated the presence of extra 2D information during both training and inference stages. To overcome the limitation of requiring extra 2D inputs and to expand the applicability of the proposed method, Semantic-Assisted Training~\cite{yang2021sat} focuses on utilizing 2D semantics during the training stage. In this paper, we aim to investigate the potential of using 2D semantics exclusively during training to assist in weakly supervised 3D visual grounding task.

\section{Method}
\begin{figure*}[t]
\centering
\includegraphics[width=1\textwidth]{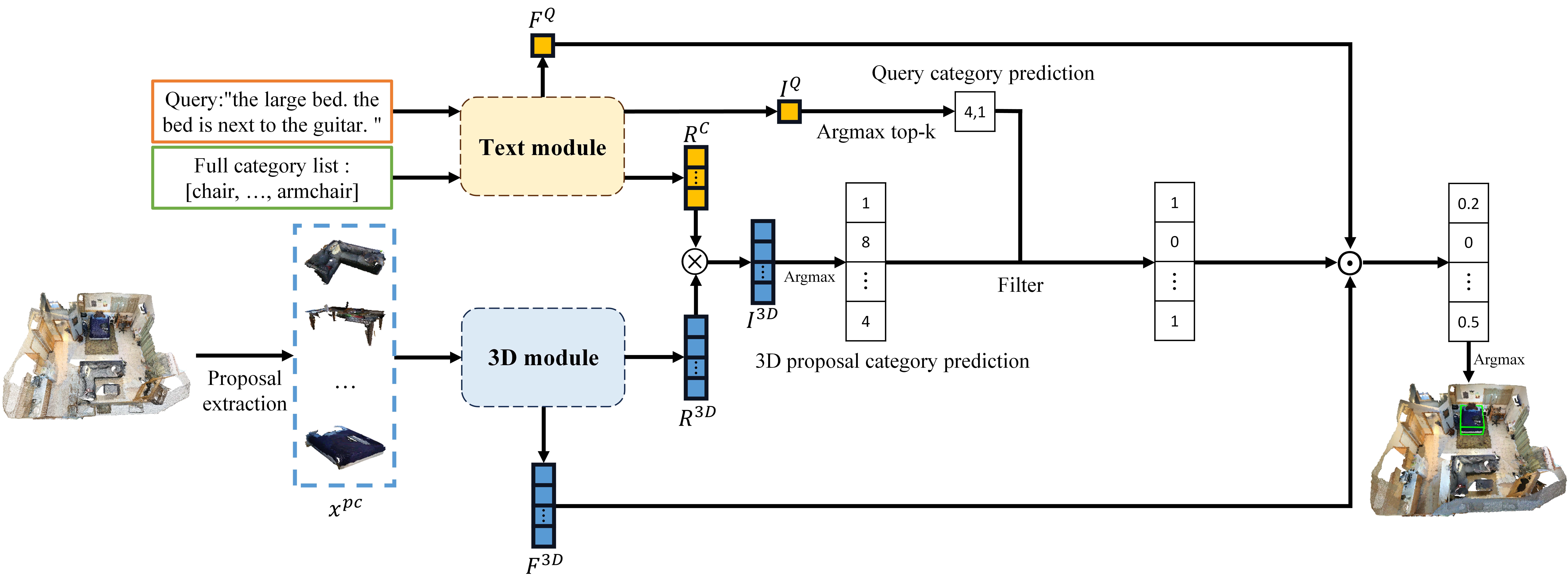} 
\caption{\small The inference procedure of our proposed 3D-VLA. Here, we only keep the text and 3D modules and does not need the 2D module. 3D proposal candidates and their embeddings ($F^{3D}$ and $R^{3D}$) are obtained from the 3D module. Text query embedding $F^{Q}$ and category label embedding $R^C$ are obtained from the text module. We perform matrix multiplication on $R^{3D}$ and $R^C$, and get the 3D proposal category prediction, and then utilize the query category prediction to filter out those proposals with different classifying results with it. For the reserved 3D proposals, we rank them by the inner product similarity between their 3D embeddings $F^{3D}$ and the text query embedding $F^Q$, and choose the top-1 proposal as the final predicted target proposal corresponding to the text query.}
\label{fig3}
\end{figure*}

In this section, we will first demonstrate our 3D-VLA training procedure by visual language alignment. Then, we will describe the inference procedure of 3D-VLA with category-oriented proposal filtering.

\subsection{3D-VLA Training by Visual Language Alignment}
As shown in Fig.~\ref{fig2}, the inputs of 3D-VLA comprises a 3D point cloud scene and a text query $Q$. The point cloud scene $S\in \mathbb{R}^{N \times 6}$, which indicates there are $N$ points in the scene, and each point is represented with six dimenstions RGB-XYZ. The 3D object proposals for the scene are readily available, either generated from the off-the-shelf 3D object detector~\cite{jiang2020pointgroup}. These proposals will serve as the initial candidate proposals for 3D visual grounding. In each dataset, the category labels are also provided for the 3D objects, we will also encode all of these category labels to get their embeddings, so as to support the coarse-grained classification task to help the model learning.

\subsubsection{3D Encoder}

For the 3D proposal candidates, we first sample 1024 points for each of them, and then leverage PointNet++~\cite{qi2017pointnet++} to do the initial feature encoding, followed by a standard transformer~\cite{vaswani2017attention} to extract higher-level 3D semantic embeddings~$F^{3D}=\left\{F_1^{3D},\dots,F_M^{3D}\right\}$, where~$M$ is the total number of 3D proposal candidates. The above procedures compose the 3D encoder $\varepsilon^{3d}$ in the 3D module.

\subsubsection{Text Encoder}
We take the text encoder of a large-scale vision-language model CLIP~\cite{radford2021learning} (other VLMs are also practicable and we choose CLIP in this paper) as the text encoder $\varepsilon^{text}$ to exact query embedding~$F^Q\in \mathbb{R}^{1\times d}$ of $Q$. Meanwhile, each category label in the full category list of the 3D visual grounding dataset is also encoded by $\varepsilon^{text}$, and represented by the category embeddings $F^C\in \mathbb{R}^{K\times d}$, where $K$ denotes the category numbers. During training, we freeze the $\varepsilon^{text}$ and directly load the CLIP pretrained parameters.

\subsubsection{2D Encoder}
\label{proj}
For each 3D proposal candidate, we project its point clouds onto $L$ sampled frames~\cite{dai2017scannet} in the original video through geometric camera calibration, and get the corresponded 2D image regions. To avoid the potential inaccuracies in the 2D-3D correspondences, we apply a boundary extension strategy after projecting the 3D point cloud onto 2D space. Specifically, we expand the projected 2D region $[x, y, w, h]$ by 10\% along both the width and height, i.e., $[x, y, w + 0.2 * w, h + 0.2 * h]$, to account for potential deviations caused by the projection. This strategy helps to capture the correct region more reliably, even when minor projection errors occur. Actually, we find that each 3D proposal may have multiple correspondences in different frames in the video and therefore refer to multiple 2D image regions. Here, we only choose the 2D image region which contains the most 3D projected points from the point cloud, to pair with the 3D proposal candidate. We leverage the image encoder of CLIP $\varepsilon^{2d}$ to extract the 2D semantic embeddings of these 2D image regions, which are denoted as~$F^{2D}=\left\{F_1^{2D}, \dots, F_M^{2D}\right\}$. Similarly, we also freeze $\varepsilon^{2d}$ and directly load the CLIP pretrained parameters.

\subsubsection{Cross-Modal Contrastive Learning}

Since large-scale vision-language models such as CLIP has established a high level of semantic alignment between 2D image embeddings and text embeddings, and we also conveniently get the 2D correspondence of each 3D point cloud proposal, we can naturally take the 2D embedding as a bridge to implicitly align the 3D embedding and text embedding with a contrastive learning process. Specifically, we follow the typical contrastive loss~\cite{khosla2020supervised} by pulling embeddings of the paired 3D proposal and 2D region closer, and pushing apart the unpaired one. The concrete definition is as follows:
\begin{equation}\label{eq1}
\begin{aligned}
\mathcal{L}_{\mathrm{e}}= & -\frac{1}{|M|} \sum_{i \in M}\left(\log \frac{\exp \left( \left(F_i^{2 \mathrm{D}} \cdot F_i^{3 \mathrm{D}}\right) / \tau\right)}{\sum_{j \in M} \exp \left( \left(F_i^{2 \mathrm{D}} \cdot {F}_j^{3 \mathrm{D}}\right) / \tau\right)}\right. \\
& \left.+\log \frac{\exp \left( \left(F_i^{2 \mathrm{D}} \cdot {F}_i^{3 \mathrm{D}}\right) / \tau\right)}{\sum_{j \in M} \exp \left( \left(F_j^{2 \mathrm{D}} \cdot {F}_i^{3 \mathrm{D}}\right) / \tau\right)}\right) .
\end{aligned} 
\end{equation}
where $\tau$ is the temperature hyper-parameter. By optimizing the $\mathcal{L}_e$ loss above, we could make the learned 3D encoder $\varepsilon^{3d}$ generate 3D proposal embedding align with its 2D image embedding, thus make it comparable to text embeddings of queries.

\subsubsection{Multi-Modal Adaption Through Task-Aware Classification}

As we known, the large-scale pretrained data of CLIP are free and general, and they do not have specialized knowledge to point cloud scenes. Therefore, only relying on the VLMs to build the 3D-text correlation will make the 3D visual grounding process not reliable. To mitigate this issue, we propose to introduce auxiliary 3D visual grounding task-aware classification to adapt the learned multi-modal embeddings better aligned in the point cloud scene.

Specifically, as shown in Fig.~\ref{fig2}, we first add an adapter each to the text, 2D and 3D modules. All these adapters are with the same structure (two fully-connected layers with ReLU activate function), and residual connections are employed to keep both the source and adapted semantics:
\begin{equation}
R^\ast=\alpha\cdot A^\ast+(1-\alpha)\cdot F^\ast,
\end{equation}
where~$\alpha$ is the ratio of residual connections. Meanwhile, to further ensure a cohesive connection between the 2D and 3D embeddings after the adaption procedure, we also introduce a contrastive loss $\mathcal L_a$ to the adapted 2D and 3D embeddings~$A_{2D}$ and~$A_{3D}$. Here, $\mathcal L_a$ fully follows $\mathcal L_e$ above and we omit its definition in this section.

Furthermore, to bring in the 3D visual grounding task-aware semantic knowledge to the overall model, we introduce three classification tasks based on the residual embeddings $R^Q$, $R^{2D}$, and $R^{3D}$. We first add a text classifier on the residual query embedding $R^Q$ to predict the distribution on the category labels of the 3D visual grounding dataset, supervised by a cross-entropy loss $\mathcal{L}^{q}_{cls}$, which we denote as query classification loss. For the 2D and 3D classification, we adopt a task-aware classification strategy. As we mentioned before, all the category labels are encoded by the text encoder $\varepsilon^{text}$, here we also input all the category embeddings to the adapter in the text module, and thus obtain the residual category embeddings $R^C$. We perform matrix multiplication on $R^C \in \mathbb{R}^{K \times d}$ and the 2D residual embeddings $R^{2D} \in \mathbb{R}^{M \times d}$, and thus get the 2D classification logits $I^{2D} \in \mathbb{R}^{M \times K}$. Softmax layer is applied on  $I^{2D}$ and a 2D classification cross-entropy loss $\mathcal{L}^{2d}_{cls}$ is introduced to supervise the above 2D image classification procedure. Symmetrically, we can also compute the 3D classification loss $\mathcal{L}^{3d}_{cls}$.

Just by introducing the above coarse-grained classification signals while without the need for fine-grained box annotations, we can make the learned adapted embeddings have better semantic awareness of the point clouds of indoor scenes, thus assisting the 3D visual grounding process.

\begin{table*}[h!]
\centering
\small
\caption{\small Performance comparison on the ScanRefer dataset.}

\setlength{\tabcolsep}{0.4mm}{
\begin{tabular}{cccccccccc}
\hline 
\multirow{2}{*}{Supervision} & \multirow{2}{*}{Method}    & \multirow{2}{*}{Pub.}   & \multirow{2}{*}{Input} & \multicolumn{2}{c}{Unique} & \multicolumn{2}{c}{Multiple} & \multicolumn{2}{c}{Overall} \\

                             &                            &                         &                        & Acc@0.25    & Acc@0.50   & Acc@0.25     & Acc@0.50    & Acc@0.25    & Acc@0.50    \\
\hline 
\multirow{8}{*}{Fully Supervised}  & ReferIt3D~\cite{achlioptas2020referit3d}                  & ECCV20                  & 3D                     & 53.75      & 37.47     & 21.03       & 12.83      & 26.44      & 16.90      \\
                             & \multirow{2}{*}{ScanRefer~\cite{chen2020scanrefer}} & \multirow{2}{*}{ECCV20} & 3D                     & 67.64      & 46.19     & 32.06       & 21.26      & 38.97      & 26.10      \\
                             &                            &                         & 3D+2D                  & 76.33      & 53.51     & 32.73       & 21.11      & 41.19      & 27.40      \\
                             & TGNN~\cite{huang2021text}                       & AAAI21                  & 3D                     & 68.61      & 56.80     & 29.84       & 23.18      & 37.37      & 29.70      \\
                             & InstanceRefer~\cite{yuan2021instancerefer}              & ICCV21                  & 3D                     & 77.45      & 66.83     & 31.27       & 24.77      & 40.23      & 32.93      \\
                             & SAT~\cite{yang2021sat}     & ICCV21                  & 3D+2D                  & 73.21      & 50.83     & 37.64       & 25.16      & 44.54      & 30.14      \\
                             & 3D-SPS~\cite{luo20223d}                     & CVPR22                  & 3D+2D                  & 84.12      & 66.72     & 40.32       & 29.82      & 48.82      & 36.98      \\
                             & EDA~\cite{wu2023eda}  & CVPR23     & 3D   & 85.76    & 68.57     & 49.13       & 37.64      & 54.59      & 42.26   \\
                             & HAM~\cite{chen2022ham}     & -   & 3D    & 79.24      & 67.86     & 41.46       & 34.03      & 48.79      & 40.60      \\
                             & M3DRef-CLIP~\cite{zhang2023multi3drefer}   & ICCV23       & 3D    & -    & \textbf{77.2}      & -    & 36.8       & -            & 44.7  \\
                             & ConcreteNet~\cite{unal2023three}         & -     & 3D    & 82.39      & 75.62     & 41.24       & 36.56      & 48.91      & 43.84      \\
                             &  3D-VisTA~\cite{zhu20233d}         &  ICCV23     &  3D    &  77.40      &  70.90     &  38.70       &  34.80      &  45.90      &  41.50      \\
                             &  G$^{3}$-LQ~\cite{wang2024g}         &  CVPR24     &  3D    &  \textbf{88.59}      &  73.28     &  \textbf{50.23}       &  \textbf{39.72}      &  \textbf{55.95}      &  \textbf{44.72}      \\
\hline 
\multirow{3}{*}{ Zero Shot} &  LERF~\cite{kerr2023lerf}         &  ICCV23     &  3D+2D    &  -      &  -     &  -       &  -      &  4.4      &     0.3      \\
&  Openscene~\cite{peng2023openscene}         &  CVPR23     &  3D+2D    &  -      &  -     &  -       &  -      &  14.3      &    4.7      \\
&  LLM-Grounder~\cite{yang2024llm}         &  ICRA24     &  3D+2D    &  -      &  -     &  -       &  -      &  \textbf{17.1}      &   \textbf{5.3}      \\

\hline 
\multirow{2}{*}{Weakly Supervised} & Wang \textit{et al}.~\cite{wang2023distilling}    & ICCV23   & 3D            & -            & -            & -            & -            & 27.37   & 21.96 \\
& Ours     & -        & 3D+2D                  & \textbf{72.95}      & \textbf{62.17}     & \textbf{22.77}       & \textbf{17.94}      & \textbf{32.51}      & \textbf{26.53}   \\
\hline 
\end{tabular}

}
\label{Tab.1}
\end{table*}

\begin{table*}[h!]
\centering
\small
\caption{\small Performance comparison on the ScanRefer dataset.}

\setlength{\tabcolsep}{0.4mm}{
\begin{tabular}{cccccccccc}
\hline 
\multirow{2}{*}{Supervision} & \multirow{2}{*}{Method}    & \multirow{2}{*}{Pub.}   & \multirow{2}{*}{Input} & \multicolumn{2}{c}{Unique} & \multicolumn{2}{c}{Multiple} & \multicolumn{2}{c}{Overall} \\

                             &                            &                         &                        & Acc@0.25    & Acc@0.50   & Acc@0.25     & Acc@0.50    & Acc@0.25    & Acc@0.50    \\
\hline 
\multirow{8}{*}{Fully Supervised}  & ReferIt3D~\cite{achlioptas2020referit3d}                  & ECCV20                  & 3D                     & 53.75      & 37.47     & 21.03       & 12.83      & 26.44      & 16.90      \\
                             & \multirow{2}{*}{ScanRefer~\cite{chen2020scanrefer}} & \multirow{2}{*}{ECCV20} & 3D                     & 67.64      & 46.19     & 32.06       & 21.26      & 38.97      & 26.10      \\
                             &                            &                         & 3D+2D                  & 76.33      & 53.51     & 32.73       & 21.11      & 41.19      & 27.40      \\
                             & TGNN~\cite{huang2021text}                       & AAAI21                  & 3D                     & 68.61      & 56.80     & 29.84       & 23.18      & 37.37      & 29.70      \\
                             & SAT~\cite{yang2021sat}     & ICCV21                  & 3D+2D                  & 73.21      & 50.83     & 37.64       & 25.16      & 44.54      & 30.14      \\
                             & 3D-SPS~\cite{luo20223d}                     & CVPR22                  & 3D+2D                  & 84.12      & 66.72     & 40.32       & 29.82      & 48.82      & 36.98      \\
                             & EDA~\cite{wu2023eda}  & CVPR23     & 3D   & 85.76    & 68.57     & 49.13       & 37.64      & 54.59      & 42.26   \\
                             & M3DRef-CLIP~\cite{zhang2023multi3drefer}   & ICCV23       & 3D    & -    & \textbf{77.2}      & -    & 36.8       & -            & 44.7  \\
                             &  G$^{3}$-LQ~\cite{wang2024g}         &  CVPR24     &  3D    &  \textbf{88.59}      &  73.28     &  \textbf{50.23}       &  \textbf{39.72}      &  \textbf{55.95}      &  \textbf{44.72}      \\
\hline 
\multirow{2}{*}{Weakly Supervised} & Wang \textit{et al}.~\cite{wang2023distilling}    & ICCV23   & 3D            & -            & -            & -            & -            & 27.37   & 21.96 \\
& Ours     & -        & 3D+2D                  & \textbf{72.95}      & \textbf{62.17}     & \textbf{22.77}       & \textbf{17.94}      & \textbf{32.51}      & \textbf{26.53}   \\
\hline 
\end{tabular}

}
\label{Tab.1}
\end{table*}

\begin{table*}[!h]
\centering
\small
\caption{\small Performance comparison on the ReferIt3D (Nr3D and Sr3D) dataset.}

\setlength{\tabcolsep}{1.0mm}{
\begin{tabular}{cccccccc}
\hline 
Supervision                 & Method        & Pub.   & Overall      & Easy         & Hard         & View-dep.    & View-indep.  \\
\hline 
\multicolumn{8}{c}{Nr3D}                                                                                                        \\
\hline 
\multirow{7}{*}{Fully Supervised} & ReferIt3D~\cite{achlioptas2020referit3d}     & ECCV20 & 35.6±0.7 & 43.6±0.8 & 27.9±0.7 & 32.5±0.7 & 37.1±0.8 \\
                            & TGNN~\cite{huang2021text}          & AAAI21 & 37.3±0.3 & 44.2±0.4 & 30.6±0.2 & 35.8±0.2 & 38.0±0.3 \\
                            & InstanceRefer~\cite{yuan2021instancerefer} & ICCV21 & 38.8±0.4 & 46.0±0.5 & 31.8±0.4 & 34.5±0.6 & 41.9±0.4 \\
                            & SAT~\cite{yang2021sat}           & ICCV21 & 49.2±0.3 & 56.3±0.5 & 42.4±0.4 & 46.9±0.3 & 50.4±0.3 \\
                            & LanguageRefer~\cite{roh2022languagerefer} & CoRL22 & 43.9       & 51.0       & 36.6       & 41.7       & 45.0       \\
                            & 3D-SPS~\cite{luo20223d}        & CVPR22 & 51.5±0.2 & 58.1±0.3 & 45.1±0.4 & \textbf{48.0±0.2} & 53.2±0.3 \\
                            & BUTD-DETR~\cite{jain2022bottom}     & ECCV22 & 54.6       & \textbf{60.7}       & 48.4       & 46.0       & \textbf{58.0}       \\
                            & EDA~\cite{wu2023eda}           & CVPR23 & 52.1       & -            & -            & -            & -            \\
                            & HAM~\cite{chen2022ham}           & -      & 48.2       & 54.3       & 41.9       & 41.5       & 51.4       \\
                        &   3D-VisTA~\cite{zhu20233d}       &   ICCV23    &   57.5      &   -       &   49.4       &   -       &   -    \\
                            &  G$^{3}$-LQ~\cite{wang2024g}     &  CVPR24      &   \textbf{58.4}       &  -       &  \textbf{50.7}       &  -       &  -     \\
\hline 
Weakly Supervised           & Ours          & -      & \textbf{32.1±0.2} & \textbf{38.6±0.2} & \textbf{25.8±0.3} & \textbf{28.8±0.3} & \textbf{33.7±0.4} \\
\hline 
\multicolumn{8}{c}{Sr3D}                                                                                                        \\
\hline 
\multirow{7}{*}{Fully Supervised} & ReferIt3D~\cite{achlioptas2020referit3d}     & ECCV20 & 40.8±0.2 & 44.7±0.1 & 31.5±0.4 & 39.2±1.0 & 40.8±0.1 \\
                            & TGNN~\cite{huang2021text}          & AAAI21 & 45.0±0.2 & 48.5±0.2 & 36.9±0.5 & 45.8±1.1 & 45.0±0.2 \\
                            & InstanceRefer~\cite{yuan2021instancerefer} & ICCV21 & 48.0±0.3 & 51.1±0.2 & 40.5±0.3 & 45.4±0.9 & 48.1±0.3 \\
                            & SAT~\cite{yang2021sat}           & ICCV21 & 57.9       & 61.2       & 50.0       & 49.2       & 58.3         \\
                            & LanguageRefer~\cite{roh2022languagerefer} & CoRL22 & 56.0       & 58.9       & 49.3       & 49.2       & 56.3       \\
                            & 3D-SPS~\cite{luo20223d}        & CVPR22 & 62.6±0.2 & 56.2±0.6 & 65.4±0.1 & 49.2±0.5 & 63.2±0.2 \\
                            & BUTD-DETR~\cite{jain2022bottom}     & ECCV22 & 67.0       & \textbf{68.6}       & 63.2       & \textbf{53.0}       & \textbf{67.6}       \\
                            & EDA~\cite{wu2023eda}           & CVPR23 & 68.1       & -            & -            & -            & -            \\
                            & HAM~\cite{chen2022ham}           & -      & 62.5       & 65.9       & 54.6       & 52.5       & 63.0       \\
                            &  3D-VisTA~\cite{zhu20233d}        &  ICCV23   &  69.6      &  -       &  63.6        &  -       &  -    \\
                            &  G$^{3}$-LQ~\cite{wang2024g}      &  CVPR24     &  \textbf{73.1}       &  -       &  \textbf{66.3}       &  -       &  -     \\
\hline 
Weakly Supervised           & Ours          & -      & \textbf{34.5±0.2} & \textbf{37.7±0.2} & \textbf{27.0±0.4} & \textbf{35.3±0.5} & \textbf{34.5±0.2} \\
\hline 
\end{tabular}
}
\label{Tab.2}
\end{table*}

\begin{table*}[]
\centering
\small
\caption{\small Performance comparison on the ReferIt3D (Nr3D and Sr3D) dataset. For the ``$R@n,IoU@m$'' metric, $n=3$ and $m\in\left\{0.25,0.5\right\}$.}
\setlength{\tabcolsep}{0.8mm}{
\begin{tabular}{cccccccccccc}
\hline 
\multirow{2}{*}{Method} & \multirow{2}{*}{Pub.} & \multicolumn{2}{c}{Easy} & \multicolumn{2}{c}{Hard} & \multicolumn{2}{c}{View-dep.} & \multicolumn{2}{c}{View-indep.} & \multicolumn{2}{c}{Overall} \\
&     & $m$=0.25      & $m$=0.5      & $m$=0.25      & $m$=0.5      & $m$=0.25         & $m$=0.5        & $m$=0.25          & $m$=0.5         & $m$=0.25        & $m$=0.5       \\
\hline 
\multicolumn{12}{c}{Nr3D} \\
\hline 
Wang \textit{et al}.~\cite{wang2023distilling}   & ICCV23   & 27.3   & 21.1  & 18.0    & 14.4 & 21.6    & 16.8     & 22.9    & 18.1  & 22.5         & 17.6       \\
Ours    &  -       & \textbf{33.3}    & \textbf{25.3}   & \textbf{24.2}     & \textbf{17.3}  & \textbf{30.3}     & \textbf{20.8}      & \textbf{31.7}     & \textbf{21.6}   & \textbf{28.7}          & \textbf{21.3}        \\
\hline 
\multicolumn{12}{c}{Sr3D}   \\
\hline 
Wang \textit{et al}.~\cite{wang2023distilling}   & ICCV23   & 29.4   & 24.9  & 21.0    & 17.5  & 20.2   & 17.2     & 27.2    & 22.9  & 26.9         & 22.7       \\
Ours     & -        & \textbf{35.2}    & \textbf{28.1}    & \textbf{25.8}    & \textbf{21.1}   & \textbf{27.3}    & \textbf{22.3}      & \textbf{33.5}      & \textbf{27.4}  & \textbf{30.5}          & \textbf{24.8}        \\
\hline 
\end{tabular}
}
\label{Tab.3}
\end{table*}

\begin{table*}[!h]
\centering
\small

\caption{\small Ablation studies of the 3D-VLA components on Nr3D.}
\setlength{\tabcolsep}{1.5mm}{
\begin{tabular}{ccccccccccc}
\hline 
    & $\mathcal{L}_e$ & $\mathcal{L}_{cls}$ & Filter & Adapter & $\mathcal{L}_a$ & Overall      & Easy         & Hard         & View-dep.    & View-indep.  \\
\hline 
(a) & \checkmark  &            &            &            &             & 17.5±0.3 & 20.9±0.4 & 14.2±0.3 & 13.5±0.4 & 19.4±0.4 \\
(b) & \checkmark  & \checkmark &            &            &             & 21.8±0.2 & 26.8±0.3 & 17.0±0.4 & 16.8±0.4 & 24.3±0.3 \\
(c) & \checkmark  & \checkmark &\checkmark  &            &             & 29.7±0.4 & 36.7±0.4 & 23.0±0.4 & 28.3±0.3 & 30.4±0.4 \\
(d) & \checkmark  & \checkmark & \checkmark & \checkmark &             & 30.8±0.3 & 38.6±0.6 & 23.4±0.2 & 28.6±0.4 & 32.0±0.3 \\
(e) & \checkmark  & \checkmark & \checkmark & \checkmark & \checkmark  & \textbf{32.1±0.2} & \textbf{38.6±0.2} & \textbf{25.8±0.3} & \textbf{28.8±0.3} &\textbf{ 33.7±0.4} \\
\hline 
\end{tabular}
}
\label{Tab.4}
\end{table*}

\begin{table*}[!t]
\centering
\begin{minipage}[t]{0.48\textwidth} 
\centering
\captionsetup{width=\linewidth} 
\caption{\small 3D-VLA performance with different $k$ in the category-oriented proposal filtering strategy on Nr3D.}
\label{Tab.5}
\setlength{\tabcolsep}{1.5mm}{
\begin{tabular}{cccccc}
\hline 
Top-$k$ & Overall      & Easy         & Hard         & View-dep.    & View-indep.  \\
\hline 
1     & 31.4±0.2 & \textbf{38.9±0.4} & 24.2±0.5 & 28.2±0.4 & 33.0±0.2 \\
2     & 31.8±0.2 & 38.5±0.4 & 25.3±0.5 & \textbf{29.7±0.4} & 32.8±0.4 \\
3     & \textbf{32.1±0.2} &38.6±0.2 & \textbf{25.8±0.3} & 28.8±0.3 & \textbf{33.7±0.4} \\
4     & 31.7±0.3 & 38.9±0.5 & 24.8±0.1 & 28.5±0.2 & 33.3±0.4 \\
\hline 
\end{tabular}
}
\end{minipage}
\hfill 
\begin{minipage}[t]{0.48\textwidth} 
\centering
\captionsetup{width=\linewidth} 
\caption{\small Performance comparison with model variants on Nr3D.}
\label{Tab.6}
\setlength{\tabcolsep}{1.5mm}{
\begin{tabular}{cccccc}
\hline
     & Easy & Hard & View-dep. & View-indep. & Overall \\ \hline
Ours & 38.6 & 25.8 & 28.8      & 33.7        & 32.1    \\
RPS.  & 2.0  & 2.0  & 1.9       & 2.0         & 2.0     \\
CBWS.  & 35.8 & 22.4 & 28.4      & 29.2        & 29.0    \\
GTS.  & \textbf{43.1} & \textbf{30.6} & \textbf{30.9}     & \textbf{39.6}        & \textbf{36.7}    \\ \hline
\end{tabular}
}
\end{minipage}
\end{table*}

\begin{table*}[!t]
\centering
\begin{minipage}[t]{0.48\textwidth} 
\centering
\captionsetup{width=\linewidth} 
\caption{\small Performance comparison with different projection strategy on Nr3D.}
\label{Tab.7}
\setlength{\tabcolsep}{0.7mm}{
\begin{tabular}{cccccc}
\hline 
 Method &  Overall      &  Easy         &  Hard         &  View-dep.    &  View-indep.  \\
\hline 
 Unmodified Projection     &  31.1 &  38.6&  23.9 &  28.9 &  32.2 \\
Boundary-Extended Projection     &  32.1 & 38.6 &  25.8 &  28.8 &  33.7 \\
\hline 
\end{tabular}
}
\end{minipage}
\hfill 
\begin{minipage}[t]{0.48\textwidth} 
\centering
\captionsetup{width=\linewidth} 
\caption{\small The runtime of three datasets.}
\label{Tab.8}
\setlength{\tabcolsep}{1.0mm}{
\begin{tabular}{cccc}
\hline
                                          &  NR3D &  SR3D & ScanRefer  \\ \hline
 Getting 2D image regions~(for a room) &  336.6s   &  362.9s  & 242.3s    \\
 Inference~(for a query)                      &  0.382s &  0.384s  & 0.397s  \\ \hline
\end{tabular}
}
\end{minipage}
\end{table*}

\subsubsection{Overall Loss Functions}

Combining the above contrastive losses $\mathcal{L}_{e}$ and $\mathcal{L}_{a}$, as well as the query, 2D and 3D classiciation losses $\mathcal{L}^q_{cls}$,  $\mathcal{L}^{2d}_{cls}$ and $\mathcal{L}^{3d}_{cls}$, our overall model is optimized by:
\begin{equation}
\mathcal{L}=\lambda_1\ast(\mathcal{L}_e+\mathcal{L}_a)\;+\lambda_2\ast\mathcal{L}_{cls}^{2d}+\lambda_3\ast\mathcal{L}_{cls}^{3d}+\lambda_4\ast\mathcal{L}_{cls}^q,
\end{equation}
where~$\lambda_*$ controls the ratio of each loss term.

\subsection{3D-VLA Inference with Category-Oriented Proposal Filtering}
In the inference stage, as shown in Fig.~\ref{fig3}, we only retain the 3D and text modules and does not need the 2D module's involvement.

Firstly, we take the 3D proposal embeddings $F^{3D}$ and its residual embeddings $R^{3D}$ from the 3D module, as well as the text query embedding $F^Q$ and the category residual embeddings $R^C$ from the text module. $I^Q$ is also computed to get the query classification result on the 3D visual grounding categories. By performing matrix multiplication on $R^{3D}$ and $R^{C}$, we can get the category prediction of each 3D proposal.  In order to make the category corresponding to the target proposal more consistent with the category corresponding to the query, we propose a category-oriented proposal filtering strategy by only keeping the 3D proposals that have the same category prediction with the top-$k$ categories of the text query. For instance, as shown in Fig.~\ref{fig3}, if the top-2 category predictions of the query are ``bed'' (id: 4) and ``sofa'' (id: 1), we only keep 3D proposals whose category prediction belonging to these two categories and create a mask, 1 for the reserved proposal and 0 for the filtered one.  Finally, for the reserved proposals, we rank them by their inner product similarity between their 3D embeddings $F^{3D}$ and the query embedding $F^Q$, and choose the proposal with the highest similarity score as the predicted target proposal.

Also noted that, if the category predictions of all 3D proposals do not match with that of the query, we keep all the proposals and do not perform the filtering strategy.

\section{Experiments}

In this section, we first present our experimental settings include datasets, evaluation metrics and our implementation details. Then, we will demonstrate our 3D-VLA results and discuss the effectiveness of each our model component.

\subsection{Experiment Settings}

\subsubsection{Dataset} We evaluate our 3D-VLA on two public and widely-used datasets \textbf{ScanRefer}~\cite{chen2020scanrefer} and \textbf{ReferIt3D}~\cite{achlioptas2020referit3d}.

The ScanRefer dataset is derived from indoor 3D scene dataset ScanNet~\cite{dai2017scannet}. It is divided into two distinct parts:  ``Unique'' and ``Multiple'', which indicate that whether the scene contains more than two distractors.

The ReferIt3D dataset is also proposed based on the ScanNet dataset. It consists of two subsets: Sr3D and Nr3D. Two distinct data splits are employed in Sr3D and Nr3D. The ``Easy'' and ``Hard'' splits are divided based on the number of distractors in the scene, and the ``View-dep.” and ``View-indep.” splits are divided based on whether the referring expression is dependent on the speaker's view.

With regard to the ReferIt3D dataset, it has provided 3D proposals as well as the category labels of them in the indoor point cloud scene. Therefore, we can directly use the provided proposals as the 3D proposal candidates, and leverage the provided category labels to provide the coarse-grained supervision signals to the model. However, for the ScanRefer dataset, it does not provide the above two terms. Therefore, we employ the pretrained PointGroup~\cite{jiang2020pointgroup} as the detector to extract the proposals as well as their category labels in advance, and then utilize the pre-extracted information to help the model training. 

\begin{figure*}[!h]
\centering
\includegraphics[scale=.4]{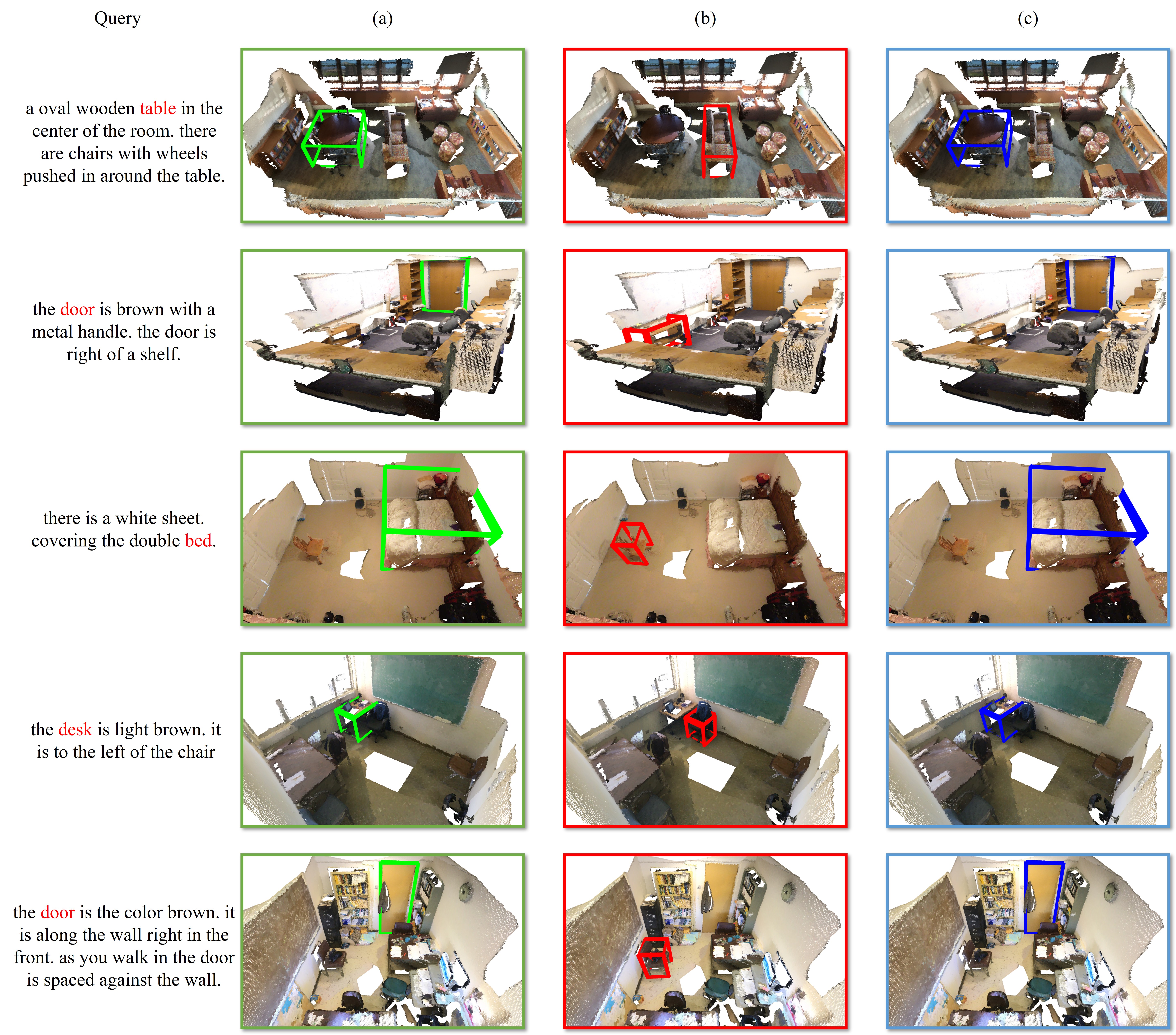} 
\caption{\small The qualitative results of our 3D-VLA on ReferIt3D dataset. We use green/red/blue colors to represent the ground truth/incorrect predictions/correct predictions. (a) shows the ground truth, (b) and (c) show our model predictions w/o and w/ category-oriented proposal filtering strategy, respectively.}
\label{fig4}
\end{figure*}

\subsubsection{Evaluation Metric}
For the ScanRefer dataset, we follow InstanceRefer~\cite{yuan2021instancerefer}, and take Acc@$m$IoU as the evaluation metric, where $m$ takes on values from the set~$\{0.25, 0.5\}$.

Since ReferIt3D dataset has provided several 3D proposals as the candidates for visual grounding, it converts the 3D visual grounding task into a classification problem, \textit{i.e.,} whether the selected proposal among the $M$ candidates is the groundtruth proposal. Models are thus evaluated by accuracy, which measures the percentage of the correct selected samples. Owing Wang \textit{et al.}~\cite{wang2023distilling} adopt their own IoU metrics on the ReferIt3D dataset, which represents the percentage of at least one of the top-$n$ predicted proposals having an IoU greater than $m$ when compared to the groundtruth target bounding box, we also follow  Wang \textit{et al.} and evaluate it on ReferIt3D dataset. Here we set $n\in3$ and $m\in\left\{0.25,0.5\right\}$.

\subsubsection{Implementations Details}

\begin{figure*}[!h]
\centering
\includegraphics[scale=.4]{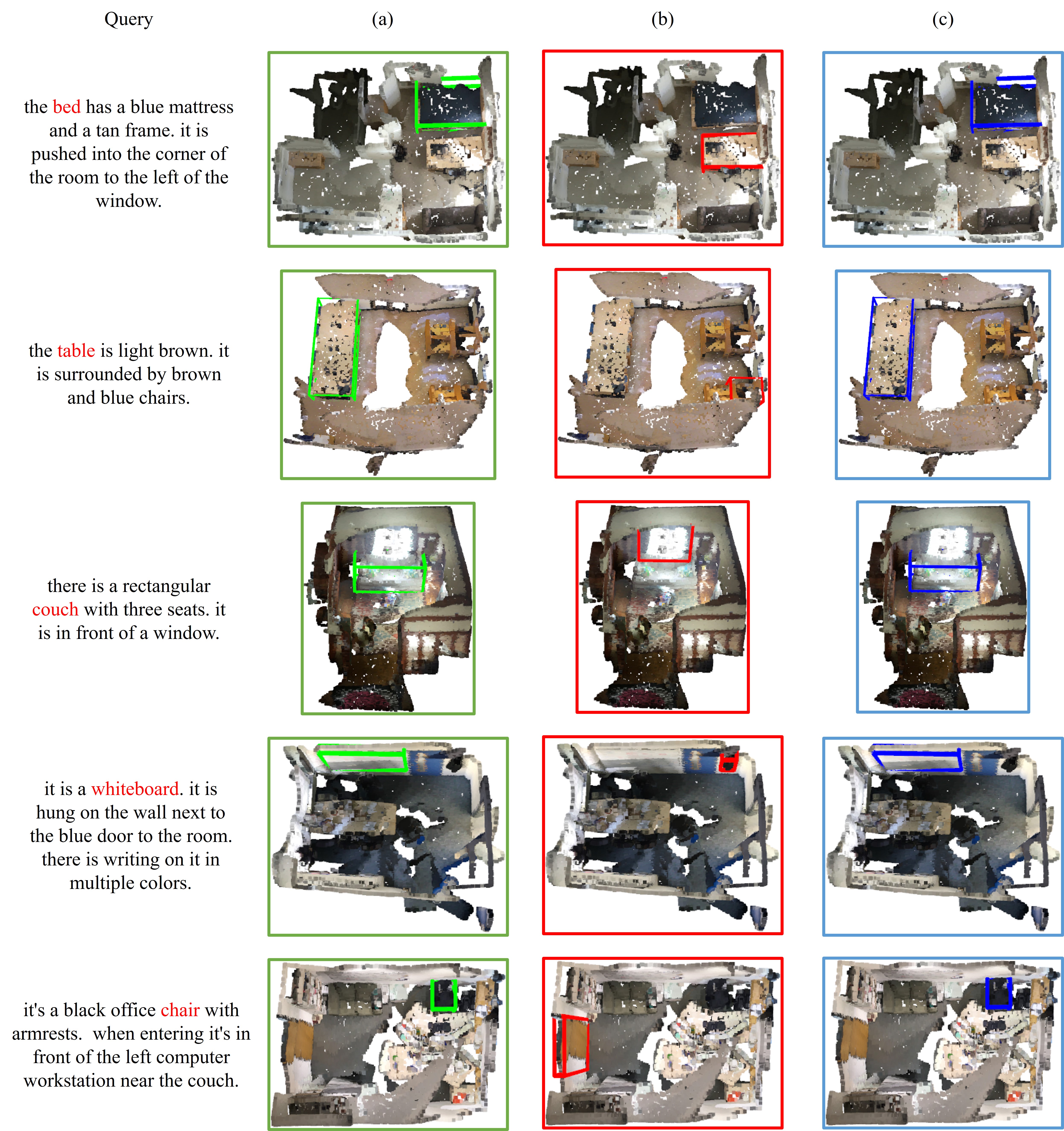} 
\caption{\small The qualitative results of our 3D-VLA on ScanRefer dataset. We use green/red/blue colors to represent the ground truth/incorrect predictions/correct predictions. (a) shows the ground truth, (b) and (c) show our model predictions w/o and w/ category-oriented proposal filtering strategy, respectively.}
\label{fig5}
\end{figure*}

3D-VLA is implemented by PyTorch~\cite{paszke2019pytorch}. Model optimization is conducted using Adam optimizer with batch size of 32. We set an initial learning rate of 0.0005 for the model, and the learning rate of the transformer layer is further adjusted by multiplying it with 0.1. We reduce the learning rate by a multiplicative factor of 0.65 at epochs 20, 30, 40, and 50. The CLIP embedding dimension $d$ is 512, and the hidden dimension in our adapters is also set as 512. Besides, we set $k = 3$ as default in category-oriented proposal filtering module.

\subsection{3D Visual Grounding Results}

\subsubsection{ScanRefer}
For the ScanRefer dataset, we present the Acc@$m$IoU performances in Table.~\ref{Tab.1}. We also indicate the used input modalities of each method (purely 3D or 3D+2D). It can be observed that, although our weakly-supervised 3D-VLA has a certain gap with the leading SOTAs of full supervised methods, we are also supervised to find that our method even outperforms some fully supervised methods. Specifically, our 3D-VLA greatly surpasses the ReferIt3D baseline~\cite{achlioptas2020referit3d} in all subsets. Furthermore, in the ``Unique'' subset, our model outperforms the ScanRefer baseline with 3D input~\cite{chen2020scanrefer} and TGNN~\cite{huang2021text} by 5.31\% and 4.34\% on Acc@0.25, and 15.98\% and 5.37\% on Acc@0.50, respectively. Although ScanRefer with 3D+2D input performs better at Acc@0.25, 3D-VLA outperforms it by a large margin on the more challenging Acc@0.50. Meanwhile, our also 3D-VLA has a 11.34\% improvement on Acc@0.50 over SAT~\cite{yang2021sat}, from 50.83\% to 62.17\% in the ``Unique'' subset. For the weakly supervised method compared, our 3D-VLA outperforms Wang \textit{et al}.~\cite{wang2023distilling} 5.14\% on Acc@0.25 and 4.57\% on Acc@0.50 and achieves the state-of-the-art performance.

We also compare our 3D-VLA with zero-shot 3D visual grounding methods. For OpenScene~\cite{peng2023openscene} and LERF~\cite{kerr2023lerf}, we follow the methodology in LLM-Grounder~\cite{yang2024llm} and apply DBSCAN clustering~\cite{ester1996density} to points with high cosine similarity between the point cloud and text embeddings. We then draw bounding boxes around these clustered points to identify target objects. Despite the zero-shot capabilities of these large models, our 3D-VLA consistently outperforms them across all subsets on ScanRefer, as shown in Table.~\ref{Tab.1}. This is primarily due to our model’s ability to leverage category information from each query, which provides prior knowledge to improve object localization. Unlike zero-shot methods, which generally lack specialization in indoor environments, our task-aware classification architecture allows the model to transfer indoor-specific embeddings, enhancing its understanding of such scenes. Additionally, our category-oriented proposal filtering isolates relevant objects by minimizing interference from irrelevant ones, further boosting the model’s localization accuracy.

\subsubsection{ReferIt3D}
In Table.~\ref{Tab.2}, we present the performance results of 3D-VLA on the ReferIt3D dataset, in comparison to the supervised models. Although our 3D-VLA does not completely outperform the supervised models across all the subsets, but the performance is still comparable. 
As shown in Table.~\ref{Tab.3}, we also follow Wang \textit{et al}.~\cite{wang2023distilling} and use their own IoU metrics on the ReferIt3D dataset. We can find that our 3D-VLA greatly surpass in all subsets compared to Wang \textit{et al}. Such results demonstrate the effectiveness and potential of our weakly supervised training diagram, which does not leverage any 3D box annotations or explicit 3D-Text correspondence supervision. 

\subsection{Ablation Studies}

\subsubsection{Effectiveness of Each Components}
In order to explore the effectiveness of the each component in our 3D-VLA, we conduct comprehensive ablation studies on the Nr3D dataset~\cite{achlioptas2020referit3d}, as shown in Table.~\ref{Tab.4}. The ablation model (a) only retains and text, 2D and 3D encoders while drops the adapters and does not use the filtering strategy. It is merely trained with the contrastive loss $\mathcal{L}_{e}$. The model (b), also does not involve adapters, but directly applies the classification losses on $F^Q$, $F^{2D}$, and  $F^{3D}$. Compared (b) to (a), we can find that introducing task-aware classification signals to guide model is beneficial to increase the 3D visual grounding accuracy. When we add the category-oriented proposal filtering in (c), the overall performance is greatly improved from 21.8\% to 29.7\%. This observation proves the effectiveness of the category-oriented proposal filtering strategy, which can filter out some confused 3D proposals with different category labels to the queries, and thus get clearer and better quality 3D proposal candidates for visual grounding. Furthermore, by introducing adapters in model (d), the performance of 3D-VLA also gets promotion. This proves that our multi-modal adaptation design can help to get a better, indoor point cloud specific embedding space to align 3D point clouds and text queries.
Finally, when introducing contrastive loss~$\mathcal{L}_a$ on the adapted embeddings, the overall model performance increases from 30.8\% to 32.1\%, and the improvements mainly come from the ``Hard'' subset and ``View-indep.'' subset. Such results show that keeping cohensive connection between the adapted embedding is beneficial for the model to identify some objects that are difficult to distinguish.

\subsubsection{Investigating the Influence of Top-$k$ Query Category Predictions in Proposal Filtering}

We investigate the influence of using different top-$k$ query category predictions in our category-oriented proposal filtering strategy. The experiments are conducted on the Nr3D dataset, and the results are shown in Table.~\ref{Tab.5}. We set $k$ in four different values, \textit{i.e.}, $k \in \{1,2,3,4\}$. It can be observed that keeping more query category predictions brings higher accuracy, which shows that keeping more possible categories from the query could provide more semantic information to filter invalid 3D proposals, and is helpful to 3D visual grounding. We take $k = 3$ as the default setting since a value of $k$ that is too large might introduce an excess of interference candidates, leading to a negative impact on the network's performance.

\subsubsection{ Investigating the influence of potential inaccuracies in the 2D-3D correspondences on overall performance}
It is well known that datasets may have potential inaccuracies in 2D-3D correspondences which is more likely to occur when the objects are small or the scene is complex. In such cases, the projected object may be too small or its location in the 2D image may be difficult to determine, leading to boundary misalignments and potential cutting errors. Additionally, misalignment of points along the object’s boundary can result in an overly small 2D area, further disrupting the localization.

Therefore, We conduct experiments using both the original, unmodified 2D projection regions and our boundary-extended approach to investigate the influence of potential inaccuracies in the 2D-3D correspondences on overall performance. The term "Unmodified Projection" refers to the method using unmodified projected areas, while the term "Boundary-Extended Projection" refers to the method using the expanded projection, as mentioned in Sec.III-A, where the projected 2D bounding box $[x, y, w, h]$ is expanded by 10\% to produce the final partition area $[x, y, w + 0.2 * w, h + 0.2 * h]$ to avoid the potential inaccuracies in the 2D-3D correspondences. The results in Table~\ref{Tab.7} show that the performance of our boundary-extended approach is comparable to that of the original 2D projection regions, indicating that the dataset's projection accuracy is generally reliable. 

\subsubsection{ Investigating the computational cost}
(a) Data Preparation Time: Prior to training, our method requires projecting the 3D point cloud to 2D to identify corresponding regions. While this projection is computationally intensive, we mitigate this by pre-computing the projections offline, significantly reducing the time burden during training (see Tab.~\ref{Tab.8} for details). (b) Computational Complexity: Our model is trained on a V100 GPU, leveraging PointNet++ as the underlying 3D architecture, which contributes to its lightweight nature. Our model requires approximately 14GB of GPU memory with batch size 48 for training and 3GB of GPU memory with batch size 48 for inference in Referit3D dataset,  demonstrating its efficiency compared to other more resource-intensive methods. (c) Inference Time: In Tab.~\ref{Tab.8}, we present the inference runtime of our method. While our model does not yet achieve real-time speeds, its inference time remains competitive, allowing for practical deployment in applications where real-time performance is not critical. We are also actively exploring optimizations to further enhance inference speed.

\subsubsection{Further Analysis}
We further analyze the performance of our model by designing several additional model variants: (a) Random Proposal Selection (RPS.): randomly selects a proposal as the target proposal for the text query; (b) CLIP-Based Weak Supervision (CBWS.): uses CLIP to compare 2D image regions and text queries, leveraging their matching results as pseudo-labels for 3D proposals and text queries; (c) Ground-Truth Supervision (GTS.): removes the 2D branch of 3D-VLA, and directly utilizes 3D labels for fully supervised training. The results are provided in Table~\ref{Tab.6}. We find that our 3D-VLA method outperforms the CLIP-Based Weak Supervision method, as pseudo-labels may be inaccurate and hinder the model's performance. Moreover, our method is more robust, leveraging natural 3D-2D correspondences for efficient embedding learning. For the Ground-Truth Supervision baseline, although our method does not always outperform it across all subsets, the performance remains comparable, which demonstrates the efficacy of our approach.

\subsubsection{Qualitative Results}

The qualitative results of 3D-VLA are shown in Fig.~\ref{fig4} and Fig.~\ref{fig5}. Compare the predictions from the column (b) to the column (c), we can find that our category-oriented proposal filtering can filter out invalid 3D proposals that have error category predictions, and thus avoid these proposals to interfere the ranking procedure of the reserved proposals.

\subsubsection{Can 3D-VLA generalize to outdoor scenes with diverse lighting and object scales?} 

While our method has been primarily evaluated on indoor datasets, we believe it has strong potential for generalization to outdoor environments. The task-aware classification architecture of our 3D-VLA is designed to adapt to diverse environmental conditions, suggesting that it should, in theory, be capable of effectively localizing target objects in outdoor scenes, irrespective of lighting variations or object scale.

\begin{figure}
\centering
\includegraphics[scale=.35]{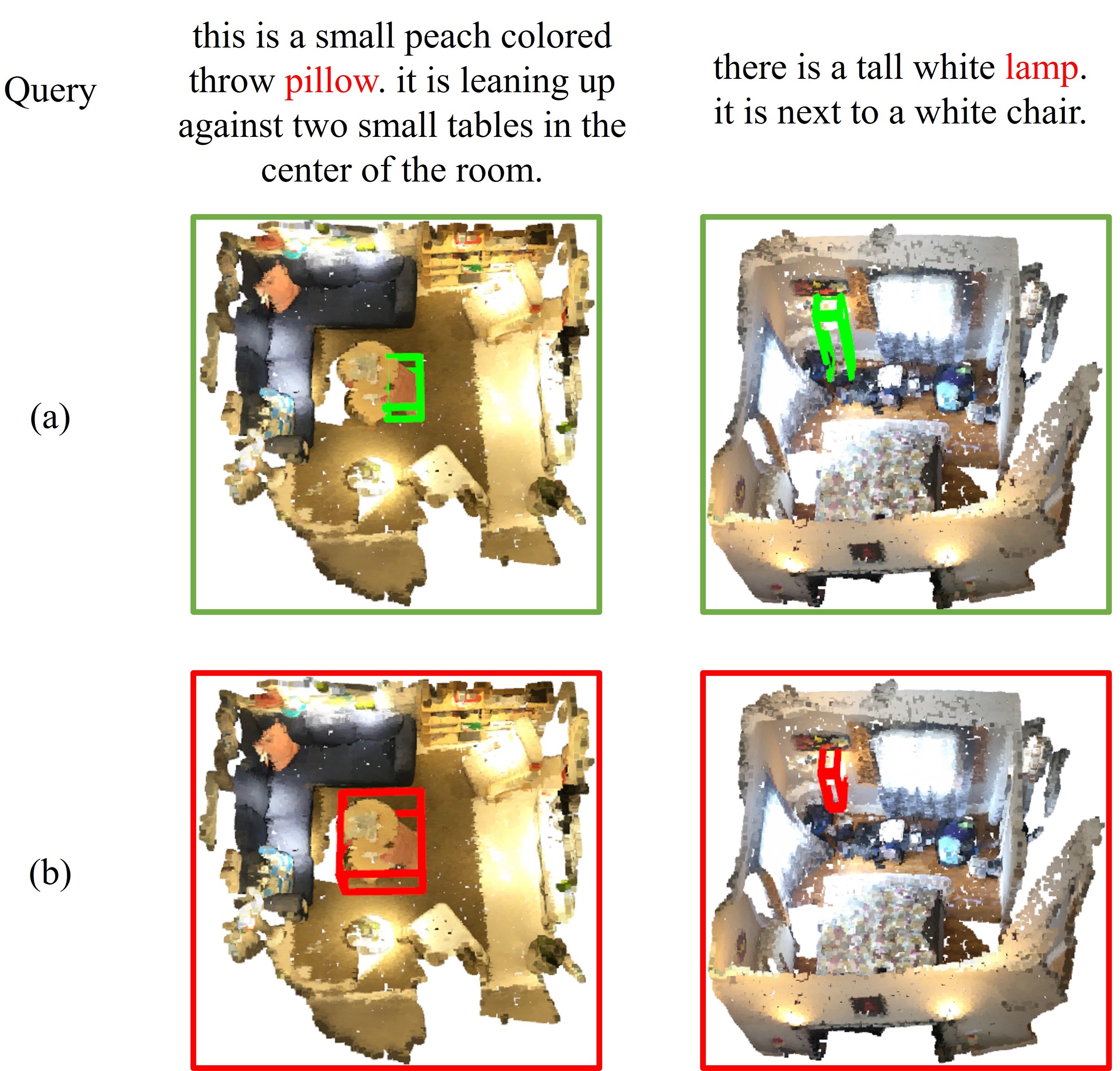} 
\caption{\small Some failure cases of our 3D-VLA. We use green/red colors to represent the ground truth/incorrect predictions. (a) and (b) show the ground truth and our model predictions, respectively.}
\label{fig6}
\end{figure}

\subsection{Limitation}
There are several limitations of our work and still much to do to realize the full potential of the proposed approach. Firstly, We still follow Wang \textit{et al}.~\cite{wang2023distilling} and employ the pretrain model to extract the proposals in advance. Therefore, as shown in Fig.~\ref{fig6}, the performance of our method is largely limited by the accuracy of the pretrained detection model. Secondly, our method still require extra 2D image during training so that it can not be applied for those datasets only with 3D point cloud. Using rendering image technology to generate high-quality 2D synthetic images may be a good solution to deal with this problem. Besides, when multiple similar objects are placed next to each other and the query involves a relation like "next," the model may struggle to disambiguate between the objects. This issue is not unique to our 3D-VLA; even fully supervised methods face challenges with ambiguous relational queries. Lastly, We recognize the potential benefits of integrating zero-shot learning techniques, especially those using large language models (LLMs) like GPT-4. Models such as LLM-Grounder leverage environmental context and relational information for better object localization. We believe incorporating these techniques into our weakly supervised framework could enhance performance. These limitation are direct avenues for future work.

\section{Conclusion}
In this paper, we propose to tackle the weakly supervised \textbf{3D} visual grounding from a novel perspective towards \textbf{V}isual \textbf{L}anguage \textbf{A}lignment, in an effort to address the shortage of object-sentence annotations. Specifically, our 3D-VLA leverages the superior ability of current advanced VLMs to align the semantics among texts and 2D images, as well as the naturally existing correspondences between 2D images and 3D point clouds, such that implicitly constructing correspondences between texts and 3D point clouds. During 3D-VLA inference, we exploit the learned text-3D correspondence to help ground the text queries to the referred 3D objects without regarding to 2D images. Through the designed scheme, a significant breakthrough is achieved than previous works, and the advantage of our 3D-VLA are also analyzed in detail. We believe these analyses can provide valuable insights to facilitate the future research of weakly supervised 3D visual grounding.

\bibliographystyle{IEEEtran}
\bibliography{IEEEabrv}

\begin{IEEEbiography}[{\includegraphics[width=1in,height=1.25in,clip,keepaspectratio]{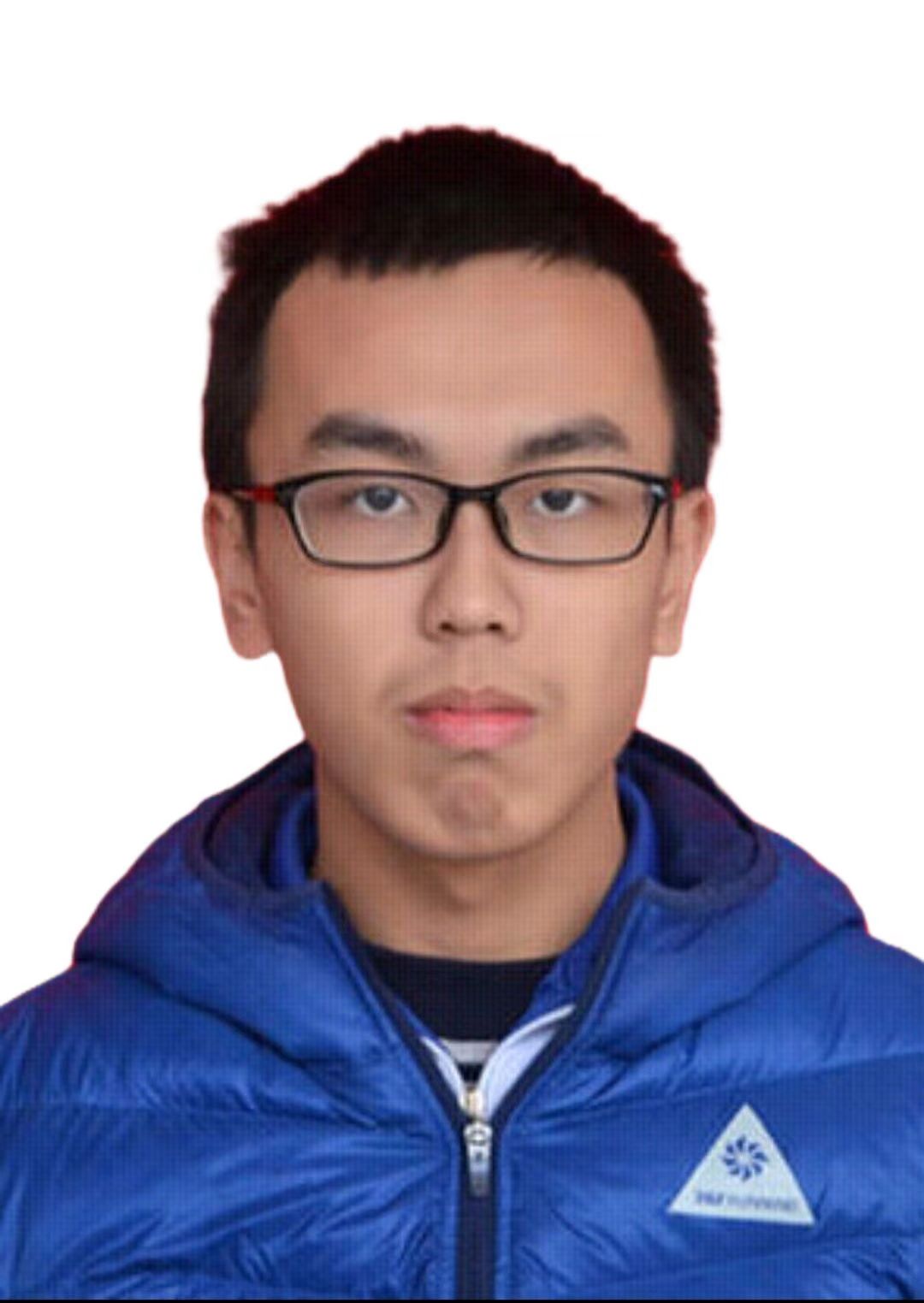}}]{Xiaoxu Xu} received the B.S. degree from the College of
Computer Science and Software Engineering, Shenzhen University, Shenzhen, China,
in 2022, and the M.S. degree from the College of
Computer Science and Software Engineering, Shenzhen University, Shenzhen, China. His research interests include 3D scene understanding and embodied AI.
\end{IEEEbiography}

\begin{IEEEbiography}[{\includegraphics[width=1in,height=1.25in,clip,keepaspectratio]{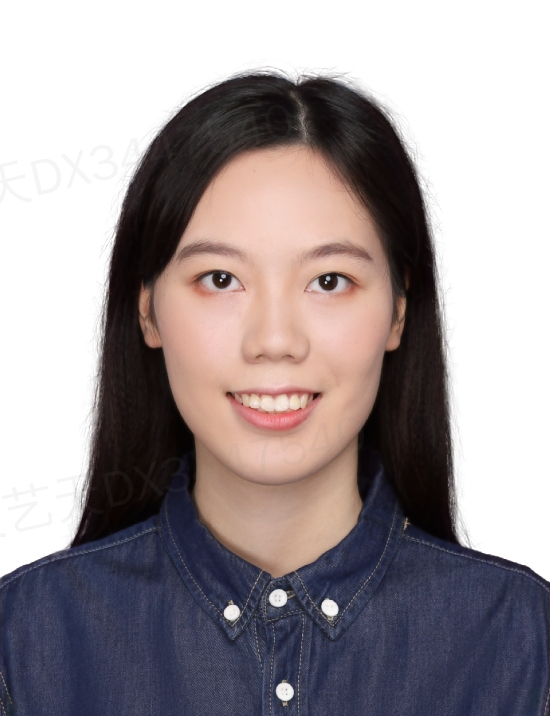}}]{Yitian Yuan} received her Ph.D. degree in computer science and technology from Tsinghua University in 2021. She got her B.E. degree in computer science from Beijing Jiaotong University in 2016. Her main research interests include computer vision and multimedia analysis, specifically for video and language, image/video/3D point cloud understanding, and multimodal LLM. She is currently a researcher in Meituan, Beijing, China.
\end{IEEEbiography}

\begin{IEEEbiography}[{\includegraphics[width=1in,height=1.25in,clip,keepaspectratio]{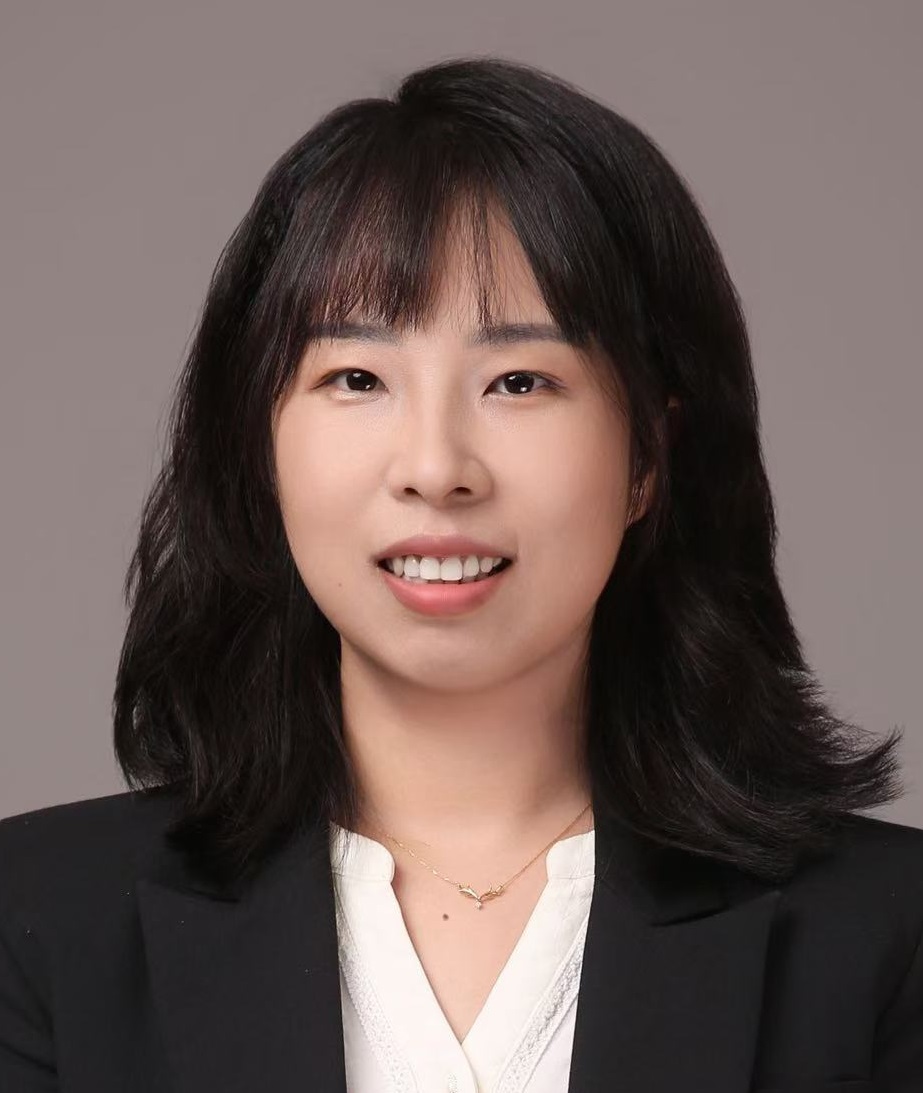}}]{Qiudan Zhang} received the B.E. and M.S. degrees in the College of Computer Science and Software Engineering from Shenzhen University, China in 2015 and 2018, respectively. She received her Ph.D. degree from the Department of Computer Science, City University of Hong Kong, China (Hong Kong SAR) in 2021. She is currently an Assistant Professor in the College of Computer Science and Software Engineering, Shenzhen University, China. Her research interests include computer vision, visual attention, 3D vision and deep learning.
\end{IEEEbiography}

\begin{IEEEbiography}
[{\includegraphics[width=1in,height=1.25in,clip,keepaspectratio]{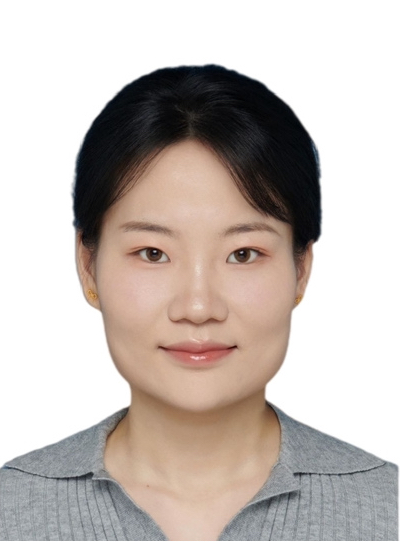}}]{Wenhui Wu} received the B.S. and M.S. degrees from Xidian University, Xian, China, in 2012 and 2015, respectively, and the Ph.D. degree in computer science from the City University of Hong Kong, Hong Kong, China, in 2019. She is currently an Associate Professor with the College of Electronics and Information Engineering, Shenzhen University, Shenzhen, China. Her current research interests include machine learning, image enhancement, and graph data analysis.
\end{IEEEbiography}

\begin{IEEEbiography}[{\includegraphics[width=1in,height=1.25in,clip,keepaspectratio]{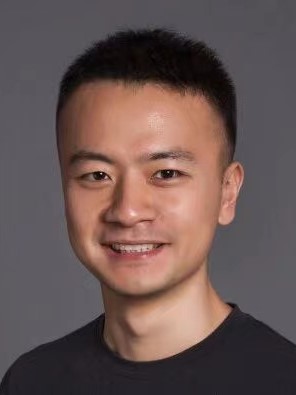}}]{Zequn Jie} received the B.E. degree from the University of Science and Technology of China, Hefei, China, and the Ph.D. degree from the National University of Singapore, Singapore. He was a Post-Doctoral Research Fellow with the Department of Electrical and Computer Engineering, National University of Singapore. He is currently a senior algorithm expert in Meituan Inc. Prior to coming to Meituan, he was a senior researcher in Tencent AI Lab. His research interests mainly fall in the fundamental computer vision topics, e.g. supervised and weakly-supervised object detection, localization and semantic segmentation. He regularly serves as a reviewer of several top-tier conferences and journals, e.g. CVPR, ICCV, ECCV, NeurIPS, ICML, TPAMI.
\end{IEEEbiography}

\begin{IEEEbiography}[{\includegraphics[width=1in,height=1.25in,clip,keepaspectratio]{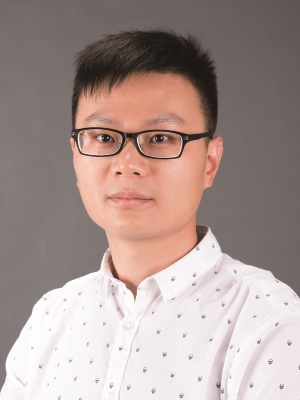}}]{Lin Ma} (M’13)  received the Ph.D. degree from the Department of Electronic Engineering, The Chinese University of Hong Kong, in 2013, the B.E. and M.E. degrees in computer science from the Harbin Institute of Technology, Harbin, China, in 2006 and 2008, respectively. He is now a Researcher with Meituan, Beijing, China.  Previously, he was a Principal Researcher with Tencent AI Laboratory, Shenzhen, China from Sept. 2016 to Jun. 2020. He was a Researcher with the Huawei Noah’Ark Laboratory, Hong Kong, from 2013 to 2016.  His current research interests lie in the areas of computer vision, multimodal deep learning, specifically for image and language, image/video understanding, and quality assessment.
		
Dr. Ma received the Best Paper Award from the Pacific-Rim Conference on Multimedia in 2008. He was a recipient of the Microsoft Research Asia Fellowship in 2011. He was a finalist in HKIS Young Scientist Award in engineering science in 2012.
\end{IEEEbiography}
	
\begin{IEEEbiography}[{\includegraphics[width=1in,height=1.25in,clip,keepaspectratio]{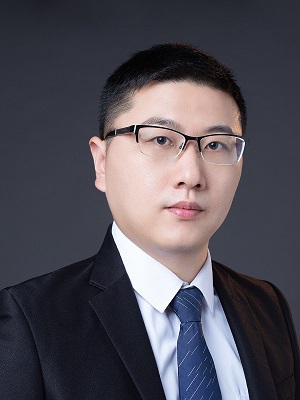}}]{Xu Wang}(M'15) received the B.S. degree from South China Normal University, Guangzhou, China, in 2007, and M.S. degree from Ningbo University, Ningbo, China, in 2010. He received his Ph.D. degree from the Department of Computer Science, City University of Hong Kong in 2014. In 2015, he joined the College of Computer Science and Software Engineering, Shenzhen University, where he is currently an Associate Professor. His research interests are video coding and 3D vision.
\end{IEEEbiography}

\vfill

\end{document}